\documentclass[letterpaper]{article} 
\usepackage[]{aaai24}  
\usepackage{times}  
\usepackage{helvet}  
\usepackage{courier}  
\usepackage[hyphens]{url}  
\usepackage{graphicx} 
\usepackage{natbib}  
\usepackage{caption} 
\usepackage{algorithm}
\usepackage{algorithmic}

\usepackage{newfloat}
\usepackage{listings}
\DeclareCaptionStyle{ruled}{labelfont=normalfont,labelsep=colon,strut=off} 
\floatstyle{ruled}
\newfloat{listing}{tb}{lst}{}
\floatname{listing}{Listing}
\nocopyright 

\usepackage{amssymb}
\usepackage{xcolor}
\usepackage{booktabs}
\usepackage[most]{tcolorbox}
\usepackage{subcaption}
\usepackage{tikz}
\usepackage{multirow}

\title{LM4OPT: Unveiling the Potential of Large Language Models in Formulating Mathematical Optimization Problems}
\author{
    Tasnim Ahmed, Salimur Choudhury
}
\affiliations{
    School of Computing, Queen's University\\

    Kingston, Ontario K7L 2N8, Canada\\
    \{tasnim.ahmed, s.choudhury\}@queensu.ca
}

\begin{document}

\maketitle

\begin{abstract}
In the rapidly evolving field of natural language processing, the translation of linguistic descriptions into mathematical formulation of optimization problems presents a formidable challenge, demanding intricate understanding and processing capabilities from Large Language Models (LLMs). This study compares prominent LLMs, including GPT-3.5, GPT-4, and Llama-2-7b, in zero-shot and one-shot settings for this task. Our findings show GPT-4's superior performance, particularly in the one-shot scenario.
A central part of this research is the introduction of `LM4OPT,' a progressive fine-tuning framework for Llama-2-7b that utilizes noisy embeddings and specialized datasets.
However, this research highlights a notable gap in the contextual understanding capabilities of smaller models such as Llama-2-7b compared to larger counterparts, especially in processing lengthy and complex input contexts.
Our empirical investigation, utilizing the NL4Opt dataset, unveils that GPT-4 surpasses the baseline performance established by previous research, achieving an F1-score of 0.63, solely based on the problem description in natural language, and without relying on any additional named entity information.
GPT-3.5 follows closely, both outperforming the fine-tuned Llama-2-7b.
These findings not only benchmark the current capabilities of LLMs in a novel application area but also lay the groundwork for future improvements in mathematical formulation of optimization problems from natural language input.
\end{abstract}

\section{Introduction}
Numerous practical challenges originating from diverse domains such as operations, economics, engineering, and computer science can be articulated as optimization problems \cite{AhmadiTeshnizi2023OptiMUSOM}. Standard optimization algorithms, including the simplex \cite{Nash2000TheS} and interior-point methods \cite{Karmarkar1984ANP}, can efficiently address these problems. Nevertheless, the translation of a real-world situation into a mathematical formulation necessitates specialized knowledge. This expertise barrier hinders many individuals from utilizing optimization algorithms, even when these could substantially enhance their operations. The advancement of automating problem formulation, which involves translating natural language descriptions into decision variables, constraints, and objective functions, has the potential to make these processes more accessible to individuals beyond just operations research experts. Consequently, optimization modeling would become accessible to individuals who cannot afford experts to augment efficiency using optimization techniques. Provided the problem is correctly formulated, it can be readily solved by transcribing it into an algebraic modeling language interpretable by solvers \cite{Ramamonjison2023NL4OptCF}.

The field of Natural Language Processing (NLP) presents a potent avenue for enhancing the accessibility and efficiency of optimization problem formulation. From the inception of word embeddings to the evolution of language models, NLP has undergone transformative progress over the years. Especially with the emergence of pre-trained language models \cite{Devlin2019BERTPO}, these models have attained state-of-the-art results on a multitude of NLP tasks such as natural language inference (NLI), question answering, summarization, collaborative writing, etc., with minimal task-specific fine-tuning \cite{Laskar2021DomainAW}. The recent advancements in LLMs, including GPT \cite{OpenAI2023GPT4TR}, and Llama \cite{Touvron2023Llama2O}, have significantly reshaped the NLP landscape and practices. These LLMs, with parameter sizes exceeding several billions, and even reaching hundreds of billions, have exhibited remarkable generalization abilities in zero-shot and few-shot settings through prompting. Furthermore, these LLMs have shown exceptional fine-tuning capabilities, even when fine-tuned on datasets significantly smaller than those used by their predecessors. 

To this end, formal assessment of this specific task$-$mathematical formulation of optimization problems from natural language descriptions using the latest developments from the GPT series models, namely GPT-3.5 and GPT-4, which have garnered widespread recognition, remains an uncharted territory. Additionally, this research aims to investigate the capabilities and limitations of a smaller Large Language Model (LLM), Llama-2-7b, when fine-tuned on this task. Consequently, this study offers the following contributions:
\begin{itemize}
\item Comprehensive analysis of GPT-3.5, GPT-4, and Llama-2-7b in mathematical formulation of optimization problems from natural language description.
\item Evaluation in zero-shot and one-shot settings to understand the impact of few-shot prompt engineering and learning adaptations of the models.
\item Empirical study using the NL4Opt \cite{Ramamonjison2023NL4OptCF} dataset, demonstrating the superior performance of GPT-4, followed by GPT-3.5.
\item Exploration of utilizing the LM4OPT framework to fine-tune Llama-2-7b, revealing significant performance enhancements.
\end{itemize}

\section{Related Work}
Efforts to simplify combinatorial optimization using LLMs have seen diverse approaches, aiming to make the process user-friendly for laypersons. The NL4Opt \cite{Ramamonjison2023NL4OptCF} competition stands out, exploring the transformation of natural language into structured optimization models. In Task 1 which is described in \cite{Dakle2023Ner4OptNE}, the aim is to accurately identify and label the components of optimization models—such as objectives, variables, and constraints—within natural language texts. Researchers approached this by using classical NER techniques that rely on the morphological and grammatical properties of the text. Additionally, modern methods were employed, involving the use of pre-trained LLMs like BERT and GPT, which were further fine-tuned on optimization-specific datasets to better understand the unique language of optimization problems. Task 2 required building mathematical representations from these elements, a more complex step involving deeper model comprehension. The methodologies here included the use of sequence-to-sequence models, which are adept at handling such translation tasks. 

The former two-step approach to generate mathematical formulation from optimization problem description requires training and dependency on two separate models. To bridge the research gap, Tsouros et al. \cite{Tsouros2023HolyG2} proposed an all-in-one LLM-based model that creates optimization models directly from prompts, showing early potential on the dataset described in NL4Opt but without established benchmarks for comparison. Advancing this approach, Teshinizi et al. \cite{AhmadiTeshnizi2023OptiMUSOM} presented a novel framework named OptiMUS, which utilizes LLMs (pre-trained GPT) to formulate and solve Mixed Integer Linear Programming (MILP) problems from natural language descriptions. They introduced a dataset, NLP4LP, containing linear programming and MILP problems to benchmark OptiMUS, which shows significant improvement over basic LLM prompting strategies. OptiMUS integrates mathematical modeling, Gurobi solver code generation, automated testing, and debugging in a cohesive system that streamlines the optimization problem-solving process. The goal of this study is to democratize access to optimization techniques across various domains, thereby broadening the use of optimization tools beyond expert circles. Furthermore, Yang et al. \cite{Yang2023LargeLM} introduced another prompt-based framework, OPRO, which uses LLMs to optimize problems without needing traditional solvers. OPRO works by iteratively improving solutions using a `meta-prompt' that incorporates both the problem description and feedback from previous solutions. It aims to learn continuously as it updates the meta-prompt with new information. To ensure stable results, OPRO generates several solutions at each iteration, balancing the need to explore different options with refining existing ones. The authors demonstrated encouraging preliminary outcomes when applying their methods to the GSM8K \cite{Cobbe2021TrainingVT} and BBH \cite{Suzgun2022ChallengingBT} datasets, in addition to tasks such as linear regression and the traveling salesman problem. The effectiveness of OPRO for complex optimization tasks is yet to be fully determined. In a recent study focused on practical applications, researchers introduced the OptiGuide framework \cite{Li2023LargeLM}, a novel integration of combinatorial optimization technology with advanced Large Language Models (LLMs), such as GPT-4, aimed at augmenting decision-making processes within supply chain management. This framework transforms user queries into in-context learning (ICL) queries for LLM processing, generating code that is vetted for accuracy and reliability. Upon validation, this code interfaces with specific components like optimization solvers and databases to derive solutions. The results, converted into understandable explanations by the LLM, simplify complex supply chain optimizations for non-technical users, fostering trust in automated decisions. In practical deployments, such as Microsoft Azure's supply chain, OptiGuide has exhibited promising outcomes, achieving an average accuracy of 93\% with GPT-4, highlighting its effectiveness in real-world settings.
A summary of the recent works in the field of Optimization and Language Models is shown in Table \ref{tab:related}.


\begin{table*}[]
\centering
\resizebox{\textwidth}{!}{%
\begin{tabular}{llllllll}
\toprule
\multicolumn{1}{c}{\textbf{Research Work}} &
  \multicolumn{1}{c}{\textbf{Dataset}} &
  \multicolumn{1}{c}{\textbf{Input}} &
  \multicolumn{4}{c}{\textbf{Framework}} &
  \multicolumn{1}{c}{\textbf{Objective}} \\
  \midrule
\multicolumn{1}{c}{} &
  \multicolumn{1}{c}{} &
  \multicolumn{1}{c}{\textbf{Problem Type in Natural Language}} &
  \multicolumn{1}{c}{\textbf{Human-in-the-loop}} &
  \multicolumn{1}{c}{\textbf{Multiple LLMs}} &
  \textbf{Fine-tuning} &
  \textbf{Prompt Engineering} &
  \multicolumn{1}{c}{} \\
  \midrule
NER4Opt &
  NL4Opt &
  Optimization &
  $\times$ &
  $\times$ &
  $\checkmark$ &
  $\times$ &
  Identifying named entitties \\
NL4Opt Competition &
  NL4Opt &
  Optimization &
  $\times$ &
  $\checkmark$ &
  $\checkmark$ &
  $\times$ &
  Mathematical Formulation \\
Holy Grail 2.0 &
  $-$ &
  Optimization &
  $-$ &
  $-$ &
  $-$ &
  $-$ &
  Mathematical Formulation \\
OPRO &
  GSM8K, BBH &
  Math word, Common-sense, Optimization &
  $\times$ &
  $\times$ &
  $\times$ &
  $\checkmark$ &
  Problem Solution \\
Optimus &
  NLP4LP &
  Optimization &
  $\checkmark$ &
  $\checkmark$ &
  $\times$ &
  $\checkmark$ &
  Problem Solution \\
Optiguide &
  \textit{Private} &
  Supply chain management &
  $\times$ &
  $\times$ &
  $\times$ &
  $\checkmark$ &
  Problem Solution (QA Session) \\
\textbf{LM4OPT (ours)} &
  \textbf{NL4Opt} &
  \textbf{Optimization} &
  \textbf{$\times$} &
  \textbf{$\times$} &
  \textbf{$\checkmark$} &
  \textbf{$\checkmark$} &
  \textbf{Mathematical Formulation}\\
  \bottomrule
\end{tabular}%
}
\caption{Recent works in the field of Optimization and Language Models}
\label{tab:related}
\end{table*}

Despite these strides, a gap persists$-$an end-to-end system that allows users the flexibility to verify and modify mathematical problem formulation, independent of the solver or programming language used. Addressing this, our research identifies a niche for benchmarking popular pre-trained LLMs on the specific task of optimization problem formulation and developing a tailored fine-tuning approach to enhance LLM specificity for this nuanced application. This work endeavors to bridge the research gap, offering a robust benchmark and a novel fine-tuning strategy that could significantly benefit the scientific community's pursuit of democratizing optimization modeling.

\section{Task Formulation}
This research investigates a generative task in the field of natural language processing, concentrating on the generation of mathematical formulations for optimization problems derived from textual descriptions. Our objective is to derive structured representations - encompassing variables, constraints, and the objective function based on given natural language descriptions. We utilize a dataset, denoted as $\mathbb{S}$, comprising a series of problem descriptions, and $\mathbf{C}$, representing their corresponding formulations in canonical mathematical form. At the core of our methodology is the introduction of an intermediate representational set, $\mathbb{R}$, which encapsulates the essential components of optimization problems (variables, constraints, and objective functions) in an equation-centric format, as opposed to the final matrix form depicted in $\mathbf{C}$. For a given problem description $s \in \mathbb{S}$, the primary goal of an LLM is to predict an intermediate representation $r \in \mathbb{R}$. Finally, the predicted intermediate representation, $r$, undergoes a systematic conversion into the canonical formulation, denoted as $c \in \mathbb{C}$, to facilitate a comprehensive evaluation of the performance of LLM. This process is exemplified in Figure \ref{fig:task}, where an example of a problem description along with the corresponding intermediate representation and canonical form is provided. It should be noted that the constraints are transformed into a format embodying `less than or equal to' conditions, and the objective function is reformulated into a minimization paradigm.

\begin{figure*}[ht!]

\begin{subfigure}{.33\textwidth}
\begin{tcolorbox}[title=Problem Description, fonttitle=\bfseries, colframe=black, colback=white, fontupper=\small, left=0mm, right=0mm, top=0mm, bottom=0mm]
A hotel employs cleaners and receptionists. Cleaners earn \$500 per week and receptionists earn \$350 per week. The hotel requires a minimum of 100 workers of whom at least 20 must be receptionists. To keep the hotel clean and running smoothly, the number of receptionists should be at least a third of the number of cleaners. The hotel wants to keep the weekly wage bill below \$30000. Formulate an LP to minimize the wage bill.
\end{tcolorbox}
\end{subfigure}%
\hfill
\begin{subfigure}{.46\textwidth}
\begin{tcolorbox}[title=Intermediate Representation, fonttitle=\bfseries, colframe=black, colback=white, fontupper=\small, left=0mm, right=0mm, top=0mm, bottom=0mm]
Variables: $cleaners, receptionists$\\ Constraints:\\ $(-1.0) * cleaners + (-1.0) * receptionists \leq -100.0$\\ $(-0.0) * cleaners + (-1.0) * receptionists \leq -20.0$\\ $(0.33) * cleaners + (-1.0) * receptionists \leq -0.0$\\ $(500.0) * cleaners + (350.0) * receptionists \leq 30000.0$\\ Objective Function:\\ $minimize (500.0) * cleaners + (350.0) * receptionist$
\end{tcolorbox}
\end{subfigure}%
\hfill
\begin{subfigure}{.19\textwidth}
\begin{tcolorbox}[title=Canonical Form, fonttitle=\bfseries, colframe=black, colback=white, fontupper=\small, left=0mm, right=0mm, top=0mm, bottom=0mm]
{[}{[}-1.0, -1.0, -100.0{]}, \\ {[}0.0, -1.0, -20.0{]}, \\ {[}0.33, -1.0, 0.0{]}, \\ {[}500.0, 350.0, 30000{]}{]}, \\ \\ {[}500.0, 350.0{]}
\end{tcolorbox}
\end{subfigure}

\caption{Task Representation}
\label{fig:task}
\end{figure*}




\section{Methodology}
In contemporary research, language models are conceptualized as functions that accept a textual input context and yield a corresponding textual output. This paradigm is predominantly instantiated through the use of transformer-based architectures, a concept introduced by Vaswani et al. \cite{Vaswani2017AttentionIA} in 2017, which has since revolutionized the field of NLP. The quintessential aspect of transformer language models is their reliance on self-attention mechanisms. These mechanisms are designed to encode input contexts by weighing the importance of different parts of the input text relative to each other. However, these models face a notable limitation in processing long text sequences due to the quadratic increase in computational complexity with longer inputs \cite{Devlin2019BERTPO}. This leads to a restricted context window during pre-training, limiting the model's ability to maintain and utilize long-term dependencies and integrate information from distant text segments. Consequently, this impacts the model’s effectiveness in tasks requiring extensive contextual understanding \cite{Brown2020LanguageMA}. To this end, our experiments investigate the performance of LLMs in zero-shot and one-shot pre-trained settings, alongside a smaller LLM, specifically fine-tuned for the task of mathematical formulation of optimization problems.

For this purpose, we evaluate GPT-3.5, GPT-4, and Llama-2-7b models. As fine-tuning is not a prerequisite for inference in these LLMs, our approach centers on the development of optimal prompt instructions for both zero-shot and one-shot settings. This development is guided by the prompt optimization techniques delineated in \cite{Yang2023LargeLM}. Additionally, to explore the impact of fine-tuning on a task-specific dataset, we selected the Llama-2-7b model, primarily due to its comparatively lower resource demands. This model was fine-tuned using the NL4Opt dataset, allowing for an in-depth analysis of fine-tuning effects on model performance within this specific context. Optimized instructions for fine-tuning, zero-shot, and one-shot prompts are provided in Figure \ref{fig:instruction}.

\begin{figure*}[ht!]
\centering

\begin{subfigure}{\textwidth}
\begin{tcolorbox}[title=Fine-tuning Instruction, fonttitle=\bfseries, colframe=black, colback=white, fontupper=\small, left=0mm, right=0mm, top=0mm, bottom=0mm]
Imagine you are a combinatorial optimization problem solver. I will give you a problem description. Your task is to find the variables, constraints, and objective functions from that description. In your response, all the constraints must be in the less than or equal to format. Your response must contain only these 3 parts: - Variables, Constraints, and Objective Function. There must be no extra strings before or after it.
\end{tcolorbox}
\end{subfigure}

\begin{subfigure}{\textwidth}
\begin{tcolorbox}[title=Zero-shot Instruction, fonttitle=\bfseries, colframe=black, colback=white, fontupper=\small, left=0mm, right=0mm, top=0mm, bottom=0mm]
Imagine you are a combinatorial optimization problem solver. I will give you a problem description. Your task is to find the variables, constraints, and objective functions from the description. I am giving you an example response format; your output should be formatted like this. Example Response:\\
``Variables: $cleaners, receptionists$\\ Constraints:\\ $(-1.0) * cleaners + (-1.0) * receptionists \leq -100.0$\\ $(-0.0) * cleaners + (-1.0) * receptionists \leq -20.0$\\ $(0.33) * cleaners + (-1.0) * receptionists \leq -0.0$\\ $(500.0) * cleaners + (350.0) * receptionists \leq 30000.0$\\ Objective Function:\\ $minimize (500.0) * cleaners + (350.0) * receptionist$".\\
Now, below is the actual problem description that you have to solve. In your response, all the constraints must be in the less than or equal to format. Your response must contain only these 3 parts: Variables, Constraints, and Objective Function. There must be no extra strings before or after it. Problem description to solve:
\end{tcolorbox}
\end{subfigure}

\begin{subfigure}{\textwidth}
\begin{tcolorbox}[title=One-shot Instruction, fonttitle=\bfseries, colframe=black, colback=white, fontupper=\small, left=0mm, right=0mm, top=0mm, bottom=0mm]
Imagine you are a combinatorial optimization problem solver. I will give you a problem description. Your task is to find the variables, constraints, and objective functions from that description. Before that, I am giving you an example problem description and response for your understanding; Your response should be formatted like this. Example Problem Description:\\
``A hotel employs cleaners and receptionists. Cleaners earn \$500 per week and receptionists earn \$350 per week. The hotel requires a minimum of 100 workers of whom at least 20 must be receptionists. To keep the hotel clean and running smoothly, the number of receptionists should be at least a third of the number of cleaners. The hotel wants to keep the weekly wage bill below \$30000. Formulate an LP to minimize the wage bill."\\ 
Example Response for the given example problem: \\ 
``Variables: $cleaners, receptionists$\\ Constraints:\\ $(-1.0) * cleaners + (-1.0) * receptionists \leq -100.0$\\ $(-0.0) * cleaners + (-1.0) * receptionists \leq -20.0$\\ $(0.33) * cleaners + (-1.0) * receptionists \leq -0.0$\\ $(500.0) * cleaners + (350.0) * receptionists \leq 30000.0$\\ Objective Function: $minimize (500.0) * cleaners + (350.0) * receptionist$".\\ Now, below is the actual problem description that you have to solve. In your response, all the constraints must be in the less than or equal to format. Your response must contain only these 3 parts: Variables, Constraints, and Objective Function. There must be no extra strings before or after it. Problem description to solve:
\end{tcolorbox}
\end{subfigure}

\caption{Instruction set for the Prompts to LLMs}
\label{fig:instruction}
\end{figure*}

\subsection{Advanced Tuning of Llama-2-7b via LM4OPT}
A progressive fine-tuning strategy was employed for the Llama-2-7b model, enabling it to initially adapt to a broader domain context related to the final task. This preliminary adaptation phase is crucial in enhancing the model's comprehension and performance capabilities. Following this, the model undergoes further fine-tuning on a specialized, task-specific dataset, where it applies the knowledge acquired in the initial phase to achieve improved performance and generalization on the target task. Prior to its fine-tuning on the NL4Opt dataset, the model was fine-tuned on GSM8K$-$a dataset comprising high-quality, linguistically diverse grade school math word problems crafted by human problem writers \cite{Cobbe2021TrainingVT}. This sequential fine-tuning approach effectively leverages the broader contextual understanding gained from GSM8K, thereby refining the model's performance on the NL4Opt tasks.

In the fine-tuning phase, a methodological approach integrating Low-Rank Adaptations (LoRA) \cite{Hu2021LoRALA} with Parameter-Efficient Fine-Tuning (PEFT) \cite{Liu2022FewShotPF} was employed. The fine-tuning process involved carefully adjusting the low-rank matrices introduced by LoRA, ensuring minimal yet strategic changes to the pre-existing weights. This method preserves the general linguistic understanding gained from pre-training while efficiently steering it toward the specialized task of mathematical problem formulation. The effectiveness of this approach is evident in the improved ability to parse and translate complex natural language descriptions into structured mathematical representations, a crucial requirement for the NL4Opt dataset. PEFT, on the other hand, extends this concept by focusing on selectively fine-tuning a small subset of the parameters. By adopting PEFT, the fine-tuning process becomes computationally less demanding and more feasible on standard hardware, while still achieving performance comparable to full-model fine-tuning. The synergy between LoRA and PEFT in fine-tuning Llama-2-7b is particularly effective in addressing the challenges of large model adaptation to specific tasks.

Furthermore, the inclusion of Noisy Embedding Instruction Fine Tuning (NEFTune) \cite{Jain2023NEFTuneNE} further augmented the fine-tuning process. NEFTune, by integrating controlled random noise into the embedding vectors during training prevents the model from overfitting to the specifics of the training dataset, such as formatting details and exact wording. Instead, it encourages the model to generate responses that are more coherent, longer, and more diverse. A detailed configuration of our experimental setup is described in the following subsection.

The incorporation of methodologies such as progressive fine-tuning, LoRA, PEFT, and NEFTune into the conventional fine-tuning framework of Large Language Models (LLMs) has notably augmented the inferential efficacy of the Llama-2-7b model. This strategic enhancement is particularly salient for a generative language model of this scale, with a parameter count of only 7 billion, especially in intricate tasks that challenge even more extensive models like GPT-3.5 and GPT-4 in their capacity to comprehend and maintain prolonged and complex contexts.

\subsection{Experimental Setup}
\label{sec:hyperparams}
The fine-tuning of the Llama-2-7b model was conducted on an NVIDIA A40 GPU, equipped with 48 GB of VRAM, over a span of 7 epochs. This process leveraged the dataset division suggested by the authors of NL4Opt \cite{Ramamonjison2023NL4OptCF}, segregating it into training, validation, and evaluation subsets. A batch size of $4$ was employed, coupled with a gradient accumulation step of $1$, and the AdamW \cite{Loshchilov2017DecoupledWD} optimizer was utilized. The initial learning rate was set at $3e-4$, with a weight decay factor of $0.001$. A random noisy embedding strength of $5$ provided the most satisfactory results during the fine-tuning process. A maximum response sequence length of $200$ was designated, under the premise that model outputs would not exceed this threshold for this specific task. Furthermore, the implementation of Gradient Checkpointing \cite{Chen2016TrainingDN} facilitated a more resource-efficient fine-tuning framework.

An additional aspect of this research involved estimating the carbon footprint associated with the fine-tuning phase, guided by the methodology proposed by Lannelongue et al. \cite{lannelongue2021green}. This analysis revealed that each fine-tuning session of the Llama-2-7b model produced approximately $23.52$ grams of $CO_2$ emissions. Notably, this finding underscores the relatively modest environmental impact of fine-tuning the model for specialized tasks.

\section{Result and Discussion}
A comprehensive assessment of various LLMs was conducted, focusing on their capability in formulating optimization problems. This evaluation was based on prompt-based zero-shot and one-shot learning experiments. The performances of these LLMs were meticulously compared against the established baseline provided by Ramamonjison et al. \cite{Ramamonjison2023NL4OptCF}, as detailed in Table \ref{tab:res}. For a consistent and objective assessment, the same scoring mechanism employed in the baseline evaluation by Ramamonjison et al. was adopted. This approach ensures a fair and direct comparison of the performance of LLMs relative to the existing benchmark in this task.

The baseline performance in Table \ref{tab:res} is derived from a fine-tuned BART \cite{Lewis2019BARTDS} model, which operates under different input conditions compared to the LLMs. While LLMs like Llama-2 and GPT receive instruction prompts and problem descriptions in natural language, the baseline BART model is also provided with named entity information extracted from the natural language problem descriptions. This additional data potentially contributes to the baseline's competitive F1-score of 0.61. The GPT-4 model, especially in the one-shot setting, outperforms others, including the baseline and GPT-3.5, with an F1-score of 0.6330. This superior performance can be attributed to GPT-4's advanced architecture and larger dataset training, as suggested by recent studies emphasizing the enhanced contextual understanding and response accuracy in more extensive models \cite{OpenAI2023GPT4TR}. Conversely, Llama-2-7b, despite being a smaller model, shows notable performance improvements in the zero-shot setting compared to one-shot, which aligns with the findings that smaller models might struggle with longer context prompts.
\begin{table}[]
\centering
\begin{tabular}{lll}
\toprule
\textbf{Language Model} & \textbf{$k$-Shot} & \textbf{F1-score} \\
\midrule
Baseline \cite{Ramamonjison2023NL4OptCF}               & -               & 0.610             \\
Llama-2-7b              & 0               & 0.1259            \\
Llama-2-7b              & 1               & 0.1022            \\
GPT-3.5                 & 0               & 0.4381            \\
GPT-3.5                 & 1               & 0.4928            \\
GPT-4                   & 0               & 0.6072            \\
\textbf{GPT-4}          & \textbf{1}      & \textbf{0.6330}  \\
\bottomrule
\end{tabular}
\caption{\textbf{Performance evaluation of LLMs for optimization problem formulation.} The best performance in terms of F1-score is highlighted in bold. GPT-3.5 (\texttt{gpt-3.5-turbo-0613}) and GPT-4 (\texttt{gpt-4-0613}) models are accessed through OpenAI api\protect\footnotemark on November 1, 2023. Llama-2-7b model is fine-tuned using the proposed LM4OPT framework.}
\label{tab:res}
\end{table}
\footnotetext{\url{https://platform.openai.com/docs/models}}

Table \ref{tab:llama} showcases the performance comparison of the Llama-2-7b model under various fine-tuning conditions. It assesses the F1-Score across different configurations, including zero-shot and one-shot settings ($k$-Shot), with and without fine-tuning, and the application of Noisy Embeddings Fine-tuning (NEFTune). Notably, progressive fine-tuning using the LM4OPT framework (P), especially in the zero-shot setting, significantly enhances the performance, achieving the highest F1-Score of 0.1259. This indicates the efficacy of progressive fine-tuning combined with NEFTune in improving the ability to understand and solve optimization problems, as opposed to non-progressive fine-tuning (N) and the baseline without any fine-tuning.

\begin{table}[]
\centering
\begin{tabular}{lllll}
\toprule
\textbf{Model} & \textbf{$k$-Shot} & \textbf{Fine-tune} & \textbf{NEFTune} & \textbf{F1-Score} \\
\midrule
           & 0 & $\times$           & $\times$     & 0.0036 \\
           & 0 & N & $\times$     & 0.0617 \\
           & 1 & N & $\times$     & 0.0581 \\
Llama-2-7b & 0 & N & $\checkmark$ & 0.0770 \\
           & 1 & N & $\checkmark$ & 0.0693 \\
           & \textbf{0} & \textbf{P}    & $\checkmark$ & \textbf{0.1259} \\
           & 1 & P     & $\checkmark$ & 0.1022 \\
\bottomrule
\end{tabular}
\caption{\textbf{Performance comparison of fine-tuned Llama-2-7b.} `N' in the `Fine-tune' column represents non-progressive fine-tuning, whereas, `P' refers to progressive finetuning. The best performance is highlighted in bold.}
\label{tab:llama}
\end{table}

A notable observation from Table \ref{tab:llama} is the superior outcomes in zero-shot settings compared to their one-shot counterparts across all configurations. This phenomenon could be attributed to the hypothesis that a smaller model like Llama-2-7b struggles with longer contexts. The data suggests that in scenarios involving extended contexts, the model tends to exhibit behavior indicative of hallucinations and produces repetitive responses that lack coherence with the broader context. Such patterns reinforce the notion that smaller models may face challenges in maintaining consistency and relevance in responses as the prompt length increases, a critical consideration in optimizing model performance for complex tasks.

\subsection{Effect of Progressive Fine-tuning}
As shown in Table \ref{tab:llama}, fine-tuning specifically for instruction processing significantly enhanced the performance of the Llama-2-7b model. Initially, the pre-trained Llama-2-7b, in both zero-shot and one-shot settings, exhibited substantial hallucination. A notable example of this was the model generating two distinct sets of variables within a single response, and its output format often did not align with the given prompt instructions, as demonstrated in Figure \ref{fig:llama_response}. However, the performance significantly improved after progressively fine-tuning the model. As it is evident from the response samples, the performance of the fine-tuned Llama-2-7b model significantly declined due to its inability to consistently maintain a specific response format. It is hypothesized that involving human evaluators or a human-in-the-loop approach for minor modifications to the outputs could significantly improve its efficiency. Such interventions could potentially bring the performance of a smaller model like Llama-2-7b closer to that of some of the larger models.

\begin{figure*}[ht!]
\hfill
\begin{subfigure}{.49\textwidth}
\begin{tcolorbox}[title=Pre-trained, fonttitle=\bfseries, colframe=black, colback=white, fontupper=\small, left=0mm, right=0mm, top=0mm, bottom=0mm]
\#\#\# Solution\\
```\\
Variables: $thin, stubby$\\
Constraints:\\
$(50.0) * thin + (90.0) * baking <= 3000.0$\\
$(30.0) * thin + (150.0) * baking <= 4000.0$\\
$(5.0) * thin + (9.0) * baking <= 0.0$\\
$(5.0) * thin + (9.0) * baking <= 0.0$\\
Objective Function: \\
$maximize (5.0) * thin + (9.0) * baking$\\
```\\
\textcolor{red}{\#\#\# Solution\\
```\\
Variables: $cleaners, receptionists$\\
Constraints:\\
$(-1.0) * cleaners + (-1.0) * receptionists <= -100.0$\\
$(-0.0) * cleaners + (-1.0$}
\end{tcolorbox}
\end{subfigure}%
\hfill
\begin{subfigure}{.49\textwidth}
\begin{tcolorbox}[title=Fine-tuned, fonttitle=\bfseries, colframe=black, colback=white, fontupper=\small, left=0mm, right=0mm, top=0mm, bottom=0mm]
Example Response: ``Variables: $thin jar, stubby jar$\\
Constraints: \\
$(50.0) * thin jar + (30.0) * stubby jar <= 3000.0$\\
$(90.0) * thin jar + (150.0) * stubby jar <= 4000.0$\\
Objective Function: \\$maximize (5.0) * thin jar + (9.0) * stubby jar$"

\end{tcolorbox}
\end{subfigure}
\caption{\textbf{Sample response from pre-trained and fine-tuned Llama-2-7b.} Hallucinating behavior is highlighted in red.}
\label{fig:llama_response}
\end{figure*}

\subsection{Does Increased Instruction Length Always Enhance Performance?}
Upon a thorough examination of the results and the outputs from both GPT and Llama models, it became evident that longer instructions do not universally enhance responses across all models. The study noted that extended, detailed instructions were beneficial for larger models like GPT-3.5 and GPT-4. Longer instructions aided GPT-3.5 and GPT-4 in resolving parsing issues common in scenarios where multiple formulations are correct but are scored differently due to manual parsing in the scoring mechanism. By specifying the solution format, these larger models were guided to respond in a particular way. For instance, GPT-3.5 in a zero-shot setting produced the equation $large_{ships} \leq small_{ships}$, whereas in a one-shot setting, it generated $large_{ships} - small_{ships} \leq 0$. The latter formulation, after parsing, yielded a higher score. However, with Llama-2-7b, a smaller model, longer instructions led to issues such as repetition and hallucination, particularly noticeable in one-shot settings. As illustrated in Figure \ref{fig:loop}, Llama-2-7b not only repeated parts of the instructions but also generated nonexistent variables such as $x1, x2, x3$, deviating from the original problem description.

\begin{figure*}[ht!]

\hfill
\begin{subfigure}{.65\textwidth}
\begin{tcolorbox}[title=Looping, fonttitle=\bfseries, colframe=black, colback=white, fontupper=\small, left=0mm, right=0mm, top=0mm, bottom=0mm]
Response:``Variables: apple, pear\\
Constraints: \\
$(-1.0) * apple + (-0.0) * pear \leq -5.0$\\
$(-0.0) * apple + (-1.0) * pear \leq -10.0$\\
$(1.0) * apple + (-0.0) * pear \leq -0.0$\\
$(1.0) * apple + (2.0) * pear \leq 0.0$\\
Objective Function: $maximize (2.0) * apple + (4.0) * pear$". \\
\textcolor{red}{\#\# Part 2: An easy LP In this part, you are given a problem description and you have to find the variables and constraints. Your response must be in the following format:}
\end{tcolorbox}
\end{subfigure}%
\hfill
\begin{subfigure}{.33\textwidth}
\begin{tcolorbox}[title=Hallucination, fonttitle=\bfseries, colframe=black, colback=white, fontupper=\small, left=0mm, right=0mm, top=0mm, bottom=0mm]
Variables: \textcolor{red}{$x1, x2, x3$}\\ 
Constraints: \\
$(-1.0) * \textcolor{red}{x1} + (-0.0) *$
\end{tcolorbox}
\end{subfigure}

\caption{\textbf{Looping Behavior and Hallucination for few-shot instructions in Llama-2-7b.} The repetitive and hallucinating behaviors are highlighted in red.}
\label{fig:loop}
\end{figure*}

\section{Limitations}
In this study, certain limitations have been identified that bear on the research outcomes. A noticeable constraint within the dataset utilized for this research is its composition of straightforward, formally structured samples replete with specific optimization domain terminologies like `formulate an LP.' This framework diverges from our overarching aim to assess the efficacy of LLMs in interpreting and formulating optimization problems as presented in natural language by individuals unversed in domain-specific jargon. It is posited that this dataset limitation might yield a discrepancy between the documented performance of LLMs and their practical application by domain-agnostic users. Moreover, resource constraints impeded the exploration of progressive fine-tuning effects on larger LLMs, such as Llama-2-70b and GPT-3.5, which may have offered additional insights. Furthermore, the adoption of a rule-based approach for converting intermediate representations to canonical forms has its drawbacks. Upon meticulous review, it was observed that some LLM-generated intermediate representations were inaccurately formatted, leading to canonical forms that diverged from the ground truth. While these discrepancies influenced the LLMs' performance metrics, it is conjectured that such nuances would be within human interpretive capabilities, suggesting that a collaborative human-model approach might counterbalance the observed performance degradation linked to format conversions. The interaction between what the model produces and how humans understand it highlights an important area for future studies. It emphasizes the need to harmonize machine precision with human judgment.

\section{Conclusion}
In this study, we undertook a comprehensive evaluation of LLMs such as GPT-3.5, GPT-4, and Llama-2-7b, focusing on their ability to translate natural language descriptions into mathematical formulation of optimization problems. The research highlights that while GPT-4 exhibits superior performance in both zero-shot and one-shot scenarios, there is a notable capability gap with smaller models like Llama-2-7b, particularly in handling complex contexts. Progressive fine-tuning of Llama-2-7b, especially with noisy embeddings and specialized datasets using our proposed LM4OPT framework, significantly enhances its performance. These findings contribute to understanding the capabilities and limitations of LLMs in a novel application area, paving the way for future improvements in the field of optimization and OR. Drawing upon the foundational research by Teshnizi et al. \cite{AhmadiTeshnizi2023OptiMUSOM}, this study lays the groundwork for future extensions, wherein the intermediate mathematical representations derived from natural language descriptions in this research could serve as precursors for LLMs to generate ILP solver code in Python.

\bibliography{aaai24}

\end{document}